\documentclass[10pt,a4paper,logo]{googledeepmind}
\usepackage{natbib}
\usepackage{multirow}
\newtheorem{definition}{Definition}

\title{DecoEvo: Score-Decoupled Co-Evolution of Solver and Rubric-Generator Skills in Text Space}
\author[1,4,*]{Jiangwang Chen}
\author[1,4,*]{Zixin Song}
\author[2,4,*]{Junlin Liu}
\author[3,4,*]{Shuaiyu Zhou}
\author[4,*]{Haiyan Wu}
\author[4]{Haihan Shi}
\author[2,4]{Chenxi Zhou}
\author[4]{Hanqing Li}
\author[4]{Xiao Yang}
\author[4]{Da Zhu}
\author[4]{Guanjun Jiang}
\author[1]{Hai Wan}
\author[1,\textdagger]{Xibin Zhao}
\affil[1]{Tsinghua University}
\affil[2]{University of Chinese Academy of Sciences}
\affil[3]{Peking University}
\affil[4]{Qwen Business Unit of Alibaba}
\renewcommand{\authornotetext}{\textsuperscript{*}Equal contribution.\quad\textsuperscript{\textdagger}Corresponding author.}

\begin{abstract}
Text-space optimization adapts large language models (LLMs) by editing external natural-language artifacts rather than model weights, so the optimized artifacts remain inspectable and the model can be treated as a black box. However, most existing text-space methods keep evaluation fixed. On open-ended tasks, this can become a bottleneck: once the solver improves on the criteria a rubric measures, omitted dimensions remain invisible to the optimization signal. Simply evolving the rubric is also unreliable when updates are selected by the current solver's score, because apparent progress can come from making the rubric easier to satisfy. We introduce \textbf{DecoEvo} (\textbf{Deco}upled Co-\textbf{Evo}lution), which co-evolves a solver skill and a rubric-generator skill under decoupled objectives without using gold rubrics during optimization. The solver skill is updated using criterion-level feedback, while the rubric-generator skill is revised through complementary audits of requirement coverage and response discrimination that are independent of aggregate solver score. This separation focuses generator updates on newly exposed solver weaknesses, reducing repeated emphasis on criteria the solver already satisfies. Under each benchmark's official evaluation, DecoEvo outperforms all compared methods across five benchmarks and three LLM backbones, yielding 2.8--5.0\% relative gains over SkillOpt in the five-benchmark average.
\end{abstract}

\begin{document}

\maketitle

\section{Introduction}
\begin{figure}[t]                                                                  
    \centering                                                                     
    \includegraphics[width=0.78\columnwidth]{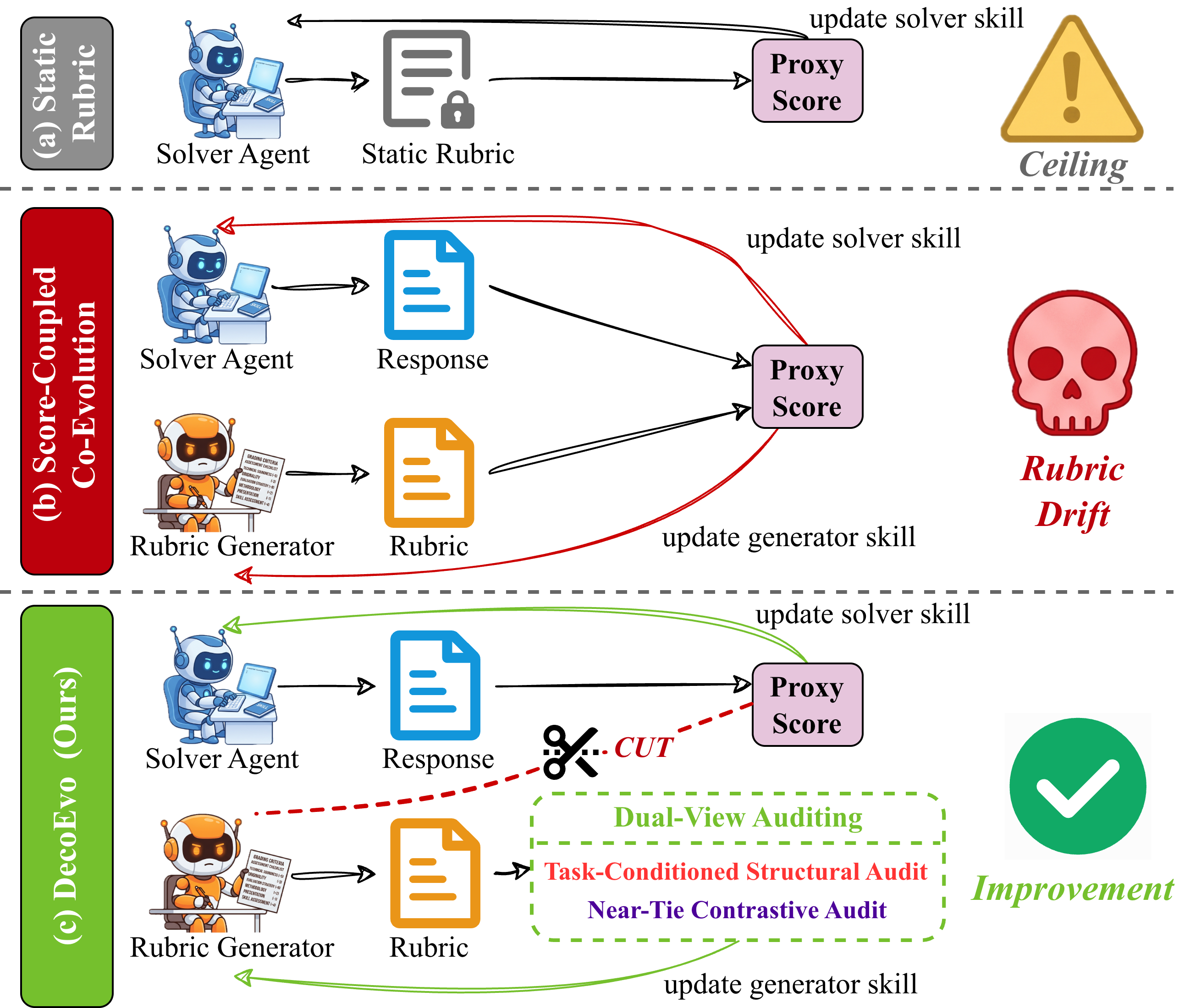}                   
    \caption{Three paradigms for rubric-based skill optimization. (a) A static rubric can bottleneck solver evolution. (b) Score-coupled co-evolution selects updates to both skills by aggregate score, risking rubric drift toward the solver. (c) DecoEvo decouples generator updates from aggregate solver score through task-conditioned structural and near-tie contrastive audits.}
    \label{fig:motivation}                                                         
\end{figure}  

Conventional LLM adaptation methods, including supervised fine-tuning, reinforcement learning from human feedback (RLHF)~\citep{ouyang2022training}, and direct preference optimization (DPO)~\citep{rafailov2023dpo}, internalize desired behavior in model parameters. Text-space optimization instead edits external natural-language artifacts, enabling black-box adaptation while keeping the optimized state inspectable.

Selecting among candidate edits still requires an evaluation objective, and existing text-space optimizers typically keep this objective fixed~\citep{skillopt,dspy,textgrad}. When the initial rubric is incomplete, a fixed objective can be limiting on open-ended tasks such as medical consultation, creative writing, and research assistance, where quality is multidimensional and multiple responses may be valid. Rubric-based LLM judges provide criterion-level feedback~\citep{geval,prometheus}, but that feedback is shaped by the dimensions specified in the rubric. As the solver addresses covered errors, the remaining failures may concentrate in dimensions that the rubric omits. For example, a medical rubric that emphasizes factual accuracy and clarity but omits contraindication warnings may not distinguish otherwise strong answers that differ in safety. The solver may then over-specialize to the measured criteria while the rubric loses discriminative value.

Updating the rubric generator can incorporate omitted criteria, but jointly adapting it with the solver creates a risk of \emph{score-coupled co-adaptation}. We represent both the task-solving strategy and the principles for generating question-specific rubrics as editable natural-language skills. If generator updates are rewarded for increasing the current solver's score, the update rule may favor criteria that the solver already satisfies over criteria that better reflect task quality. The proxy can then rise without a corresponding gain in externally evaluated task quality, causing the solver and generator to co-adapt to the proxy. This risk is related to instability in classical co-evolutionary systems~\citep{coevo_cycling} and reward overoptimization~\citep{gao2022rewardoveropt}.

Recent work adapts rubrics alongside policies in weight-space training~\citep{onlinerubrics,evolm,evorubric,evorubrics,arco}. We study a frozen black-box setting in which both solver and rubric generator are persistent text artifacts. Here, generator edits change the solver's training proxy, creating an identification problem: aggregate score cannot distinguish solver improvement from a rubric that is easier to satisfy.

We propose \textbf{DecoEvo}, a score-decoupled framework for co-evolving solver and rubric-generator skills under a frozen backbone. Generated rubrics provide criterion-level feedback to the solver; generator revisions address what current rubrics fail to cover or distinguish. A task-conditioned structural audit identifies omitted requirements, while rubric-blind comparisons of near-tied responses expose missed quality distinctions. Their diagnoses are distilled into reusable generator-skill revisions, so evidence from individual examples can improve future rubrics. This removes the direct incentive for generator updates to favor the current solver while retaining black-box compatibility and inspectability. Figure~\ref{fig:motivation} contrasts this design with static-rubric optimization and Score-Coupled Co-Evolution. Our contributions are as follows:
\begin{itemize}
    \item We formulate text-space solver--rubric co-evolution as a score-coupling problem and introduce separate update objectives for persistent solver and rubric-generator skills, allowing the evaluation signal to adapt without rewarding rubric changes that favor the current solver.
    \item We operationalize this separation with task-conditioned structural and rubric-blind near-tie audits that convert omitted requirements and missed distinctions into reusable rubric-generation principles, requiring neither gold-rubric supervision nor model weight updates.
    \item Across five benchmarks and three backbones, including two within-domain cross-benchmark transfers, DecoEvo obtains the highest mean score among the compared methods in all 15 backbone--benchmark combinations, with average relative improvements of 2.8--5.0\% over SkillOpt.
\end{itemize}

\section{Related Work}

\paragraph{Text-Space Optimization.}
Text-space optimization adapts frozen LLMs by editing external
natural-language artifacts. OPRO searches over
prompts~\citep{opro}, DSPy optimizes instructions and
demonstrations~\citep{dspy}, and TextGrad uses textual
feedback to guide revisions~\citep{textgrad}. SkillOpt is
closest on the solver side, treating a reusable skill document
as trainable state under a fixed evaluation
objective~\citep{skillopt}. Multi-agent protocol distillation
transfers structured search behavior into model
parameters~\citep{liu2026mapd}. DecoEvo additionally optimizes a
persistent rubric-generator skill under a separate acceptance
objective.

\paragraph{LLM Judges and Adaptive Rubrics.}
Recent benchmarks expose persistent weaknesses in both
specialized mathematical and general
reasoning~\citep{liu2026amobench,liu2026general365}.
Rubric-based LLM judges provide criterion-level feedback for
open-ended evaluation~\citep{geval,judgelm,prometheus}.
Contrastive rubric-generation methods derive discriminative
criteria from differences between candidate
responses~\citep{onlinerubrics,openrubrics}, while OpenRS
instantiates pairwise adaptive rubrics and refines an explicit
meta-rubric~\citep{openrs}. DR Tulu maintains evolving
rubrics during weight-space policy training for deep research
agents~\citep{drtulu}. Concurrent work explores
policy--rubric co-evolution in weight space through on-policy
or adversarial training~\citep{arco,evorubrics}. DecoEvo
co-evolves persistent text-space skills under a frozen backbone
and selects generator revisions through separate coverage and
discrimination objectives; aggregate solver score plays no role
in generator acceptance.

\paragraph{Co-Evolution and Proxy Optimization.}
Optimizing an imperfect proxy can increase measured reward without improving externally evaluated quality~\citep{gao2022rewardoveropt}. Classical co-evolutionary systems can likewise become unstable when interacting components adapt to each other rather than an external objective~\citep{coevo_cycling}. If driven by the solver's aggregate score, generator updates might similarly favor easier-to-satisfy rubrics over true task quality. The score-decoupled update removes this direct incentive while retaining response-level scores for near-tie selection and blind-spot verification.

\section{Method}

\begin{figure*}[t]
    \centering
    \includegraphics[width=\textwidth]{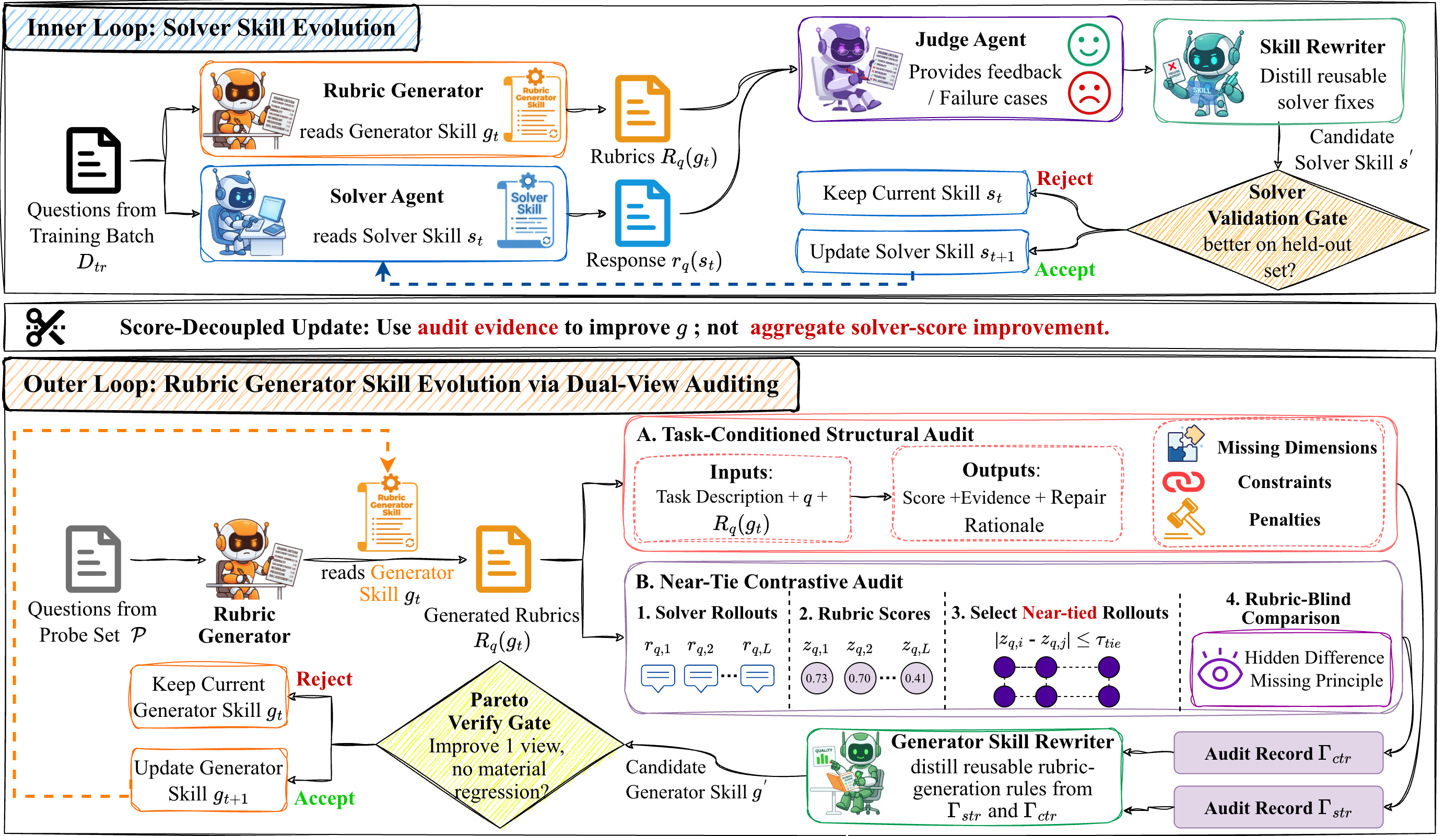}
    \caption{Overview of DecoEvo. The inner loop updates the solver skill $s_t$ from criterion-level rubric feedback. The outer loop updates the rubric-generator skill $g_t$ through two frozen, role-separated audits: a task-conditioned structural audit and a near-tie contrastive audit. Improvements in aggregate solver score under generated rubrics are never used as the generator objective.}
    \label{fig:pipeline}
\end{figure*}

\paragraph{Overview.}
DecoEvo maintains two editable natural-language skills while keeping model parameters fixed. The \emph{solver skill} $s$ stores reusable task strategies, whereas the \emph{rubric-generator skill} $g$, hereafter the \emph{generator skill}, stores principles for constructing question-specific rubrics. The solver is updated from criterion-level feedback; the generator is updated from structural and contrastive audit diagnoses distilled into reusable revisions selected on held-out audit objectives. Response-level rubric scores locate near-ties, but aggregate solver score never rewards generator updates. Thus, DecoEvo evolves a reusable generator skill rather than individual rubrics. At deployment, only the optimized solver skill is retained. Figure~\ref{fig:pipeline} summarizes the framework.

We use \emph{gold rubric} as an operational term for the fixed benchmark-specific rubric, checklist, or set of evaluation criteria held out from optimization and used only for final or post-hoc evaluation. Gold rubrics are never exposed during optimization. All optimization roles, including the solver, rubric generator, criterion-level judge, auditors, and skill rewriters, use the \emph{same frozen backbone} under fixed role-specific prompts. Thus, role separation changes the prompt and update objective, not the underlying model.

\subsection{Problem Setup}
\label{sec:setup}

Let $\mathcal{Q}$ denote a set of open-ended questions. Given $q \in \mathcal{Q}$, the solver produces a response $r_q(s)=\textsc{Solve}(q;s)$, and the generator produces a question-specific rubric
\begin{equation}
    R_q(g) = \textsc{Gen}(q; g) = \{(c_i,w_i)\}_{i=1}^{m_q},
\end{equation}
where $m_q\geq1$, $c_i$ is a natural-language criterion, and $w_i \in \mathbb{R}_{>0}$ is its point value. Training criteria are written as positively weighted requirements; undesirable behaviors are represented by avoidance requirements. This convention is independent of gold-rubric scoring rules, which may include signed criteria. A fixed criterion-level judge $E$ returns $e(q,r,c_i) \in [0,1]$. The rubric score is
\begin{equation}
    S(q,r,R_q) = \frac{\sum_i w_i e(q,r,c_i)}{\sum_i w_i},
\end{equation}
and the training-time proxy on a question set $D \subseteq \mathcal{Q}$ is
\begin{equation}
    J_D(s,g) = \frac{1}{|D|}\sum_{q\in D} S\big(q,r_q(s),R_q(g)\big).
    \label{eq:proxy_objective}
\end{equation}
The target quantity is the solver's test utility $U(s)$, computed by the benchmark's official grader using held-out gold rubrics. We use mutually disjoint data: a training pool $D_{\mathrm{tr}}$ for solver-skill edits, a probe set $\mathcal{P}$ for producing generator-audit diagnoses, and held-out sets $\mathcal{V}_s$ and $\mathcal{V}_g$ for accepting solver and generator candidates, respectively.

A fixed generator may eventually stop exposing useful new failure modes. Yet using $J_D(s,g)$ to guide generator updates is unsafe: under joint adaptation, it can rise either because the solver improves or because the generator produces easier rubrics. DecoEvo therefore separates the objectives used by the two updates.

\begin{definition}[Score-Decoupled Generator Update]
\label{def:decoupling}
Given fixed task-conditioned structural and near-tie contrastive audit roles, a generator update is \emph{score-decoupled} if its proposal and acceptance may use questions, generated rubrics, response-level scores, and structured audit records, but neither uses $J_D(s_t,g_t)$, $J_D(s_t,g')$, or their difference as the generator objective.
\end{definition}

Score decoupling retains response-level scores for locating near-ties and audit scores for verifying diagnosed repairs; aggregate solver score does not indicate generator improvement.

\subsection{Solver Skill Evolution}
\label{sec:solver}

At inner-loop step $t$, the solver answers a batch $\mathcal{B}_t \subset D_{\mathrm{tr}}$ with skill $s_t$, and the generator produces a rubric $R_q(g_t)$ for each question. The judge scores each response against every rubric criterion and returns the score breakdown and textual failure rationales. Selected failures are passed to a solver-skill rewriter, which consolidates their reusable diagnoses into a candidate skill $s'$ rather than appending question-specific fixes. For a margin $\varepsilon>0$, the candidate is accepted only when
\begin{equation}
    J_{\mathcal{V}_s}(s',g_t) > J_{\mathcal{V}_s}(s_t,g_t) + \varepsilon.
    \label{eq:solver_gate}
\end{equation}
The comparison is paired on the same validation questions and cached per-question rubrics, isolating the solver-skill edit from rubric-generation noise. Otherwise, the current skill is retained and a stall counter is incremented, providing an adaptive trigger for auditing the rubric generator.

\subsection{Generator Skill Evolution via Dual-View Auditing}
\label{sec:generator}

The outer loop runs every $K$ inner steps or when $M$ consecutive solver candidates are rejected. It follows an \emph{audit--distill--verify} procedure. First, the two frozen audit roles diagnose the current generator. Second, a generator-skill rewriter receives the audit records and distills local findings into reusable principles. Third, the candidate generator is evaluated on held-out audit cases and accepted only when it improves at least one available audit objective without materially degrading any other. This Pareto-style rule avoids aggregating heterogeneous audit scores into a single scalar.

\paragraph{Task-Conditioned Structural Audit.}
For each task family $\tau$, we define a structural-audit specification $\pi_{\mathrm{str}}^{\tau}$ using only its public task description and evaluation instructions. It serves as a \emph{task-level prior} rather than per-example expert supervision; it is fixed before optimization and never uses question-specific gold rubrics.

Given a generator skill $g$ and question $q$, the structural auditor receives $\pi_{\mathrm{str}}^{\tau}$, $q$, and the generated rubric $R_q(g)$, but no solver response or solver score. It returns an adequacy score $u_q(g)\in[0,1]$, an evidence-grounded rationale $\xi_q$, and a repair suggestion $\kappa_q$. The rationale must cite concrete question or rubric content and explain the identified omission or ambiguity; the repair suggestion states how the rubric-generation policy should change. Applying this audit to $g_t$ over $q\in\mathcal{P}$ yields the structural audit records $\Gamma_{\mathrm{str}}^t$.
Structural adequacy on a question set $D$ is
\begin{equation}
    \phi_{\mathrm{str}}(g;D)
    = \frac{1}{|D|}\sum_{q\in D}u_q(g).
    \label{eq:structural_signal}
\end{equation}
The scalar is used only for paired candidate verification; the rationale and repair suggestion provide the edit direction to the generator-skill rewriter.

\paragraph{Near-Tie Contrastive Audit.}
The near-tie contrastive audit probes what the current rubric fails to distinguish in practice. For each probe question $q$, the current solver samples $L$ stochastic rollouts $r_{q,1}^t,\ldots,r_{q,L}^t$ using $s_t$. The current rubric assigns each rollout the score $z_{q,\ell}^t=S\!\left(q,r_{q,\ell}^t,R_q(g_t)\right)$.
These scores are used only to locate pairs that the current rubric may fail to distinguish. For a threshold $0\leq\tau_{\mathrm{tie}}\leq1$, we construct
\begin{equation}
    \mathcal{N}_q^t
    = \bigl\{(i,j):1\leq i<j\leq L,\;
    |z_{q,i}^t-z_{q,j}^t|\leq\tau_{\mathrm{tie}}\bigr\},
    \label{eq:near_ties}
\end{equation}
and retain a subset $\widetilde{\mathcal{N}}_q^t\subseteq\mathcal{N}_q^t$ of at most $H$ pairs. We prioritize small score gaps and avoid repeatedly pairing the same rollout, so questions with many near-ties do not dominate the signal.

For each selected pair, the auditor follows a two-stage protocol designed to reduce anchoring to the rubric under inspection. It first sees only the question and two responses, without the rubric or scores, and returns a rubric-blind label $y_{qij}\in\{i\succ j,j\succ i,\mathrm{equivalent}\}$ together with a rationale $\xi_{qij}$ describing the task-relevant quality difference. Only after fixing this preference does it receive the generated rubric and criterion-level score breakdown. It then explains why the rubric misses the difference and proposes a missing or underweighted principle $\kappa_{qij}$. For every non-equivalent pair, we reorder the responses according to $y_{qij}$ and denote the auditor-preferred response by $r_{qij}^{+}$ and the other response by $r_{qij}^{-}$. We collect the resulting tuples $(q,r_{qij}^{+},r_{qij}^{-},\xi_{qij},\kappa_{qij})$ as the contrastive audit records $\Gamma_{\mathrm{ctr}}^t$.
Pairs judged equivalent are discarded. If $\Gamma_{\mathrm{ctr}}^t$ is empty, the contrastive audit contributes no edit request at that outer step.

For a retained record $p=(q,r^+,r^-,\xi,\kappa)$, define its rubric-induced margin as
\begin{equation}
    m_g(p)
    = S\!\left(q,r^+,R_q(g)\right)
      - S\!\left(q,r^-,R_q(g)\right).
    \label{eq:pair_margin}
\end{equation}
Given a frozen blind-spot set $\Gamma$, candidate verification uses the question-balanced resolution rate
\begin{equation}
\begin{aligned}
    \phi_{\mathrm{ctr}}(g;\Gamma)
    &= \frac{1}{|\mathcal{Q}_{\Gamma}|}
       \sum_{q\in\mathcal{Q}_{\Gamma}}
       \phi_q(g;\Gamma_q),\\
    \phi_q(g;\Gamma_q)
    &= \frac{1}{|\Gamma_q|}
       \sum_{p\in\Gamma_q}
       \mathbf{1}\!\left[m_g(p)\geq\gamma\right].
\end{aligned}
    \label{eq:contrastive_signal}
\end{equation}
Here $\Gamma_q$ contains the audited records for $q$, $\mathcal{Q}_{\Gamma}$ contains questions with at least one retained record, and $0<\gamma\leq1$ is the required resolution margin. The score is defined only when $|\Gamma|>0$. Unlike aggregate solver score, it tests whether a candidate generator resolves specific distinctions identified by the frozen auditor.

\paragraph{Audit Distillation and Pareto Verification.}
The two audits expose complementary failure modes: the structural audit asks \emph{what the rubric omits}, whereas the contrastive audit asks \emph{what the rubric fails to distinguish among score-similar responses}. Given the incumbent skill and both audit records, the generator-skill rewriter produces a candidate $g'$. It consolidates recurring question-level findings into general rules for constructing future rubrics rather than appending local patches.

Proposal and acceptance use disjoint data: the rewriter sees only probe records, whereas candidates are evaluated on $\mathcal{V}_g$, whose records remain hidden from both rewriters. For contrastive verification, the rollouts, selected pairs, rubric-blind preferences, and rationales are frozen so that $g_t$ and $g'$ are scored on the same audited set, denoted by $\Gamma_{\mathrm{ctr}}^{t,\mathcal{V}_g}$. Let $\mathcal{K}_t\subseteq\{\mathrm{str},\mathrm{ctr}\}$ denote the available audit objectives. Define
\begin{equation}
\begin{aligned}
    \Phi_{\mathrm{str}}(g)
    &= \phi_{\mathrm{str}}(g;\mathcal{V}_g),\\
    \Phi_{\mathrm{ctr}}(g)
    &= \phi_{\mathrm{ctr}}
       \bigl(g;\Gamma_{\mathrm{ctr}}^{t,\mathcal{V}_g}\bigr).
\end{aligned}
\end{equation}
The second quantity is used only when the frozen contrastive set is non-empty. Let $\Delta_k=\Phi_k(g')-\Phi_k(g_t)$ for $k\in\mathcal{K}_t$. For an improvement margin $\delta>0$ and regression tolerance $\eta\geq0$, the candidate is accepted when
\begin{equation}
    \exists k\in\mathcal{K}_t:\ \Delta_k > \delta,
    \qquad
    \forall j\in\mathcal{K}_t:\ \Delta_j \geq -\eta.
    \label{eq:gen_gate}
\end{equation}
Thus at least one available audit objective must improve by a non-trivial margin, while no available objective may regress beyond the tolerance $\eta$. No weighted cross-objective aggregation is required. If accepted, $g'$ becomes the generator skill for subsequent inner-loop steps; otherwise, the incumbent skill is retained.

Because $\mathcal{V}_s$ and $\mathcal{V}_g$ are queried by multiple accept/reject decisions, both are model-selection sets rather than unbiased evaluation sets. Their scores are never reported as final performance, and validation records are not exposed to the rewriters. Final results are computed on a disjoint test set that is not queried during solver or generator evolution. Training-trajectory analyses use an additional diagnostic set excluded from every update and selection decision.

\section{Experiments}
\label{sec:experiments}

\subsection{Experimental Setup}

\paragraph{Tasks and evaluation.} We optimize on \textbf{HealthBench}~\citep{healthbench}, \textbf{WritingBench}~\citep{writingbench}, and \textbf{ResearchQA}~\citep{researchqa}, and test direct within-domain cross-benchmark transfer from HealthBench to \textbf{LLMEval-Med}~\citep{llmeval_med} and from WritingBench to \textbf{EQ-Bench Creative Writing v3}~\citep{creative_writing_bench_v3}. The transfer questions are never used for optimization or tuning. We test \textbf{GPT-4o}~\citep{gpt4o_system_card} and \textbf{Qwen3-4B/8B}~\citep{qwen3}; within each setting, the same frozen backbone fills every optimization role. At final evaluation, responses are scored against gold rubrics using each task's official grader, prompt, and scalar metric. These graders are separate from the training-time criterion-level judge and never propose or select updates. Metrics are linearly mapped to $0$--$100$ where needed; no common grader or metric is substituted.

\paragraph{Baselines.} \textbf{Zero-shot} performs no optimization. \textbf{SkillOpt} optimizes the same persistent solver-skill representation as DecoEvo but keeps the initial rubric generator fixed. Our matched \textbf{Score-Coupled Co-Evolution (SC-CoEvo)} baseline holds DecoEvo's initialization, solver updates, proposal-side audits, candidate proposals, trigger rule, and validation splits fixed, but accepts generator edits using the current solver's aggregate score under generated rubrics. Together they isolate the progression from no optimization, to solver-only optimization, to score-coupled joint adaptation; targeted alternative-explanation controls are introduced with their results below.

\paragraph{Protocol and optimization.} Each source task uses disjoint pilot, training, probe, solver-gate, generator-gate, diagnostic, and test partitions. The pilot selects hyperparameters, only probe diagnoses reach the generator rewriter, the gates select candidates, and the diagnostic set supports only post-hoc trajectories. The final source-optimized solver skill is applied to the corresponding transfer benchmark without further updates. Audit schedules, rollout counts, and optimization thresholds are provided in the supplementary material.

\paragraph{Reporting.} We run five matched end-to-end seeds with otherwise fixed settings. The main table reports mean $\pm$ sample standard deviation, while compact secondary tables report five-run means. Paired 95\% confidence intervals are reported below for the primary comparisons, with additional intervals provided in the supplementary material. Avg. is computed within each run as the unweighted mean over the five benchmarks and then aggregated across runs.

\subsection{Main Results}

\begin{table*}[t]
    \centering
    {\small
    \setlength{\tabcolsep}{3.5pt}
    \begin{tabular}{ll | ccccc | c}
    \toprule
    \textbf{Model} & \textbf{Method} & \textbf{HealthBench} & \textbf{LLMEval-Med}$^\dagger$ & \textbf{WritingBench} & \textbf{Creative Writing}$^\dagger$ & \textbf{ResearchQA} & \textbf{Avg.} \\
    \midrule
    \multirow{4}{*}{GPT-4o}
    & Zero-shot & 40.9$\pm$0.28 & 56.5$\pm$0.36 & 62.8$\pm$0.31 & 80.0$\pm$0.44 & 65.6$\pm$0.39 & 61.2$\pm$0.18 \\
    & SkillOpt & 41.7$\pm$0.35 & 57.6$\pm$0.33 & 63.9$\pm$0.29 & 80.5$\pm$0.31 & 64.8$\pm$0.47 & 61.7$\pm$0.10 \\
    & SC-CoEvo & 40.3$\pm$0.52 & 53.0$\pm$0.61 & 63.5$\pm$0.45 & 79.2$\pm$0.55 & 65.1$\pm$0.50 & 60.2$\pm$0.16 \\
    & \textbf{DecoEvo} & \textbf{45.3}$\pm$0.23 & \textbf{60.2}$\pm$0.30 & \textbf{66.1}$\pm$0.27 & \textbf{82.7}$\pm$0.28 & \textbf{69.7}$\pm$0.34 & \textbf{64.8}$\pm$0.13 \\
    \midrule
    \multicolumn{2}{c|}{Relative gain (\%)} & 8.6\%$\uparrow$ & 4.5\%$\uparrow$ & 3.4\%$\uparrow$ & 2.7\%$\uparrow$ & 7.6\%$\uparrow$ & 5.0\%$\uparrow$ \\
    \midrule
    \multirow{4}{*}{Qwen3-4B}
    & Zero-shot & 37.6$\pm$0.41 & 63.6$\pm$0.32 & 60.1$\pm$0.49 & 70.2$\pm$0.57 & 63.8$\pm$0.43 & 59.1$\pm$0.33 \\
    & SkillOpt & 39.9$\pm$0.46 & 64.2$\pm$0.38 & 61.1$\pm$0.41 & 71.0$\pm$0.48 & 65.8$\pm$0.51 & 60.4$\pm$0.24 \\
    & SC-CoEvo & 38.9$\pm$0.58 & 63.4$\pm$0.53 & 61.0$\pm$0.49 & 69.2$\pm$0.62 & 65.3$\pm$0.59 & 59.6$\pm$0.24 \\
    & \textbf{DecoEvo} & \textbf{41.5}$\pm$0.24 & \textbf{65.1}$\pm$0.27 & \textbf{62.9}$\pm$0.34 & \textbf{73.8}$\pm$0.36 & \textbf{67.4}$\pm$0.33 & \textbf{62.1}$\pm$0.19 \\
    \midrule
    \multicolumn{2}{c|}{Relative gain (\%)} & 4.0\%$\uparrow$ & 1.4\%$\uparrow$ & 2.9\%$\uparrow$ & 3.9\%$\uparrow$ & 2.4\%$\uparrow$ & 2.9\%$\uparrow$ \\
    \midrule
    \multirow{4}{*}{Qwen3-8B}
    & Zero-shot & 40.8$\pm$0.38 & 65.0$\pm$0.30 & 60.2$\pm$0.42 & 74.9$\pm$0.51 & 69.0$\pm$0.41 & 62.0$\pm$0.31 \\
    & SkillOpt & 43.0$\pm$0.36 & 65.6$\pm$0.35 & 61.8$\pm$0.37 & 75.2$\pm$0.40 & 70.4$\pm$0.39 & 63.2$\pm$0.15 \\
    & SC-CoEvo & 41.5$\pm$0.48 & 64.4$\pm$0.52 & 60.4$\pm$0.47 & 75.4$\pm$0.54 & 69.4$\pm$0.50 & 62.2$\pm$0.24 \\
    & \textbf{DecoEvo} & \textbf{45.2}$\pm$0.24 & \textbf{66.9}$\pm$0.28 & \textbf{63.5}$\pm$0.29 & \textbf{77.7}$\pm$0.35 & \textbf{71.6}$\pm$0.27 & \textbf{65.0}$\pm$0.09 \\
    \midrule
    \multicolumn{2}{c|}{Relative gain (\%)} & 5.1\%$\uparrow$ & 2.0\%$\uparrow$ & 2.8\%$\uparrow$ & 3.3\%$\uparrow$ & 1.7\%$\uparrow$ & 2.8\%$\uparrow$ \\
    \bottomrule
    \end{tabular}
    }
    \caption{Main results under gold rubrics. $^\dagger$ marks direct within-domain cross-benchmark transfer. Avg. is the unweighted five-benchmark mean; entries are mean $\pm$ sample standard deviation over five runs. SkillOpt keeps the rubric generator fixed; SC-CoEvo denotes Score-Coupled Co-Evolution. Relative gain is measured against SkillOpt.}
    \label{tab:main_results}
\end{table*}

Table~\ref{tab:main_results} establishes effectiveness; subsequent controls, trajectories, and rubric-alignment analyses test whether task priors, additional compute, or more permissive rubrics explain the gains.

\paragraph{Overall effectiveness.} DecoEvo leads all baselines on all 15 backbone--benchmark pairs. Relative to SkillOpt, its average gains are 5.0\%, 2.9\%, and 2.8\% for GPT-4o, Qwen3-4B, and Qwen3-8B, corresponding to 3.10, 1.74, and 1.78 points. SkillOpt itself exceeds Zero-shot in 14 of 15 cases, so the solver can benefit from a fixed rubric generator initially; the remaining gap suggests that incomplete rubric coverage may remain a bottleneck after solver-only optimization.

\paragraph{Cross-benchmark transfer and score coupling.} Without any updates on the transfer benchmarks, DecoEvo improves over SkillOpt by 1.4--4.5\% on LLMEval-Med and 2.7--3.9\% on Creative Writing. The transfer gains are consistent with the optimized solver skill encoding reusable strategies shaped by broader evaluation feedback rather than source-question corrections. Conversely, SC-CoEvo falls below SkillOpt in 13 of 15 cases and below Zero-shot in seven. This reversal highlights the generator acceptance rule: aggregate-score selection can erase solver-only gains. The trajectory study below directly examines how its score-coupled selection signal diverges from the fixed external standard.

\paragraph{Paired analysis.} Across five matched runs, average gains over SkillOpt are 3.10 points for GPT-4o (95\% confidence interval $[2.88,3.32]$), 1.74 for Qwen3-4B ($[1.42,2.06]$), and 1.78 for Qwen3-8B ($[1.69,1.87]$); these prespecified backbone-level comparisons all remain significant after Holm correction~\citep{holm1979}.

\subsection{Ablation Study}

\begin{table}[t]
    \centering
    {\small
    \setlength{\tabcolsep}{5.5pt}
    \begin{tabular}{lccc}
    \toprule
    \textbf{Variant} & \textbf{GPT-4o} & \textbf{Qwen3-4B} & \textbf{Qwen3-8B} \\
    \midrule
    DecoEvo & 45.3 & 41.5 & 45.2 \\
    w/o Structural Audit & 43.6 & 40.2 & 43.7 \\
    w/o Contrastive Audit & 43.2 & 39.8 & 43.3 \\
    w/o Rubric Blindness & 44.0 & 40.5 & 44.0 \\
    w/o Near-Tie Sampling & 43.8 & 40.3 & 43.8 \\
    w/o Verification & 42.4 & 39.1 & 42.5 \\
    w/o Distillation & 43.5 & 40.1 & 43.6 \\
    \bottomrule
    \end{tabular}
    }
    \caption{Ablation study on HealthBench, averaged over five matched runs.}
    \label{tab:ablation}
\end{table}

Table~\ref{tab:ablation} ablates one mechanism at a time on HealthBench across all three backbones: either audit, rubric-blind preference formation, near-tie pair selection, held-out verification, or cross-example distillation. Removing an audit leaves the generator gate on the remaining objective; all variants otherwise share the same splits, schedules, hyperparameters, and matched seeds.

Both audits matter: removing contrastive auditing causes slightly larger losses (1.7--2.1 points) than removing structural auditing (1.3--1.7). On HealthBench, this suggests that after broad task dimensions are covered, the remaining bottleneck lies in distinctions among plausible outputs. Rubric blindness and near-tie sampling contribute 1.0--1.5 points, showing that both pair construction and blind judgment matter. Removing verification causes the largest loss (2.4--2.9 points), as unchecked generator edits can corrupt later feedback. Removing distillation costs 1.4--1.8 points, indicating that reusable evaluation principles outperform episodic suggestions. Full definitions and paired intervals for the smallest effect are in the supplementary material.

\subsection{Alternative-Explanation Controls}

\begin{table}[b]
    \centering
    {\small
    \setlength{\tabcolsep}{8pt}
    \begin{tabular}{lccc}
    \toprule
    \textbf{Method} & \textbf{GPT-4o} & \textbf{Qwen3-4B} & \textbf{Qwen3-8B} \\
    \midrule
    SkillOpt & 41.7 & 39.9 & 43.0 \\
    Task-Prior & 42.4 & 40.3 & 43.4 \\
    Budget-Matched & 42.1 & 40.4 & 43.6 \\
    Direct Audit & 43.7 & 40.7 & 44.1 \\
    DecoEvo & \textbf{45.3} & \textbf{41.5} & \textbf{45.2} \\
    \bottomrule
    \end{tabular}
    }
    \caption{Controls on HealthBench, averaged over five matched runs. SkillOpt keeps the rubric generator fixed.}
    \label{tab:main_controls}
\end{table}

The gains could instead come from richer task priors, more optimization calls, or audit feedback reaching the solver directly. Table~\ref{tab:main_controls} tests these alternatives on HealthBench under the same splits, final grader, and matched seeds. Task-Prior adds the structural-audit specification to a fixed rubric-generator skill but receives no online records; Budget-Matched spends additional solver updates toward paired DecoEvo call and token caps; Direct Audit routes the same proposal-side audit records directly to the solver rewriter. Table~\ref{tab:cost} reports realized usage.

\begin{table}[t]
    \centering
    {\small
    \setlength{\tabcolsep}{3.9pt}
    \begin{tabular}{lrrrr}
    \toprule
    \textbf{Method} & \textbf{Calls (k)} & \textbf{Tokens (M)} & \textbf{Token Rel.} & \textbf{Score} \\
    \midrule
    SkillOpt & 4.8 & 22.1 & 1.00$\times$ & 41.7 \\
    Budget-Matched & 9.1 & 41.7 & 1.89$\times$ & 42.1 \\
    Direct Audit & 9.0 & 41.2 & 1.86$\times$ & 43.7 \\
    SC-CoEvo & 9.1 & 41.7 & 1.89$\times$ & 40.3 \\
    DecoEvo & 9.1 & 41.7 & 1.89$\times$ & \textbf{45.3} \\
    \bottomrule
    \end{tabular}
    }
    \caption{Mean per-run optimization cost on HealthBench with GPT-4o, excluding the one-time pilot search and common final evaluation. SkillOpt keeps the rubric generator fixed; SC-CoEvo denotes Score-Coupled Co-Evolution. Token Rel. normalizes token use to SkillOpt; only the final solver is deployed.}
    \label{tab:cost}
\end{table}

Task priors improve SkillOpt by only 0.4--0.7 points, and budget matching by 0.4--0.6, so neither richer public information nor additional compute alone accounts for DecoEvo's 1.6--3.6-point gain. Direct Audit recovers a larger portion, indicating that the audit records carry useful information about solver failures, but remains 0.8--1.6 points below DecoEvo. This residual gap suggests that where the audit signal is stored matters: an evolving generator skill turns episodic diagnoses into persistent criteria that reshape future solver feedback, whereas direct routing affects only the current solver rewrite. Paired intervals are reported in the supplementary material.

\subsection{Training Dynamics}
\label{sec:dynamics}

\begin{figure*}[t]
    \centering
    \includegraphics[width=\textwidth]{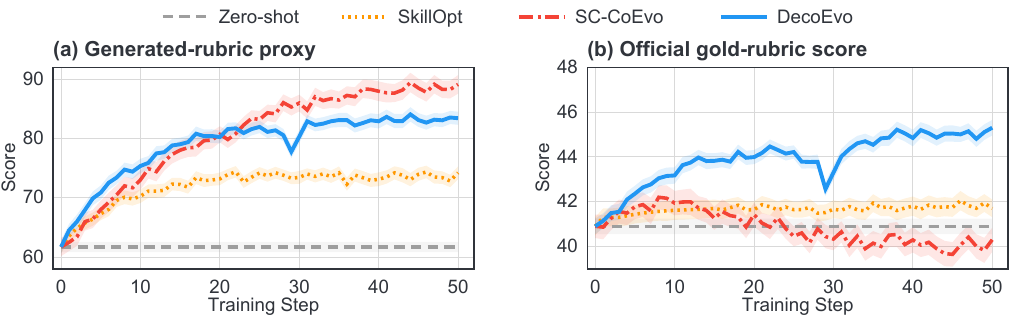}
    \caption{Training dynamics on a held-out HealthBench diagnostic set. Generated-rubric proxies use the GPT-4o training-time judge; gold-rubric scores use the official grader. Lines show five-run means with 95\% confidence bands.}
    \label{fig:training_dynamics}
\end{figure*}

Figure~\ref{fig:training_dynamics} tracks every saved HealthBench checkpoint on a fixed diagnostic set. At each checkpoint, the same responses are scored by the GPT-4o training-time judge under the generated rubric and by the official HealthBench grader under the gold rubric. Neither diagnostic-set score is used for optimization. Because the generated rubric changes during co-evolution, its proxy is not calibrated across methods or checkpoints, and changes in proxy magnitude are therefore not interpreted as changes in task quality. The comparable signal is the official gold-rubric trajectory: SC-CoEvo peaks and then declines despite retaining a high internal proxy of 89.1 at the final checkpoint. This pattern illustrates that its internal selection signal can remain favorable even as performance under the fixed external standard deteriorates. SkillOpt eventually plateaus, whereas DecoEvo recovers from transient fluctuations and finishes with the highest official gold-rubric score.

\subsection{Efficiency and Optimization Cost}

Table~\ref{tab:cost} compares methods that are informative about optimization cost on HealthBench with GPT-4o. Zero-shot has no optimization phase, while Task-Prior does not materially change the SkillOpt budget, so they are omitted here. Calls and input-plus-output tokens are averaged over five runs and exclude the one-time pilot search and common final evaluation; deployment retains only the final solver skill.

Budget-Matched, SC-CoEvo, and DecoEvo all round to 1.89$\times$ SkillOpt's token use, yet respectively gain 0.4, lose 1.4, and gain 3.6 points. Thus, realized compute alone does not explain performance. Direct Audit recovers 2.0 of the 3.6 points with 98.8\% of DecoEvo's tokens, confirming that the audits carry useful signal while indicating that persistent generator evolution explains much of the remaining gain. Since the generator is discarded after optimization, this tradeoff affects development cost rather than deployment latency.

\subsection{Rubric Alignment}

\begin{table}[t]
    \centering
    {\small
    \setlength{\tabcolsep}{2.8pt}
    \begin{tabular}{l ccc ccc ccc}
    \toprule
    & \multicolumn{3}{c}{\textbf{HealthBench}} & \multicolumn{3}{c}{\textbf{WritingBench}} & \multicolumn{3}{c}{\textbf{ResearchQA}} \\
    \cmidrule(lr){2-4} \cmidrule(lr){5-7} \cmidrule(lr){8-10}
    \textbf{Method} & P & R & F1 & P & R & F1 & P & R & F1 \\
    \midrule
    SkillOpt & 54.2 & 37.8 & 44.6 & 57.6 & 41.3 & 48.1 & 55.9 & 39.7 & 46.4 \\
    SC-CoEvo & 47.5 & 32.1 & 38.3 & 51.8 & 36.5 & 42.8 & 49.3 & 34.8 & 40.8 \\
    DecoEvo  & \textbf{63.7} & \textbf{51.4} & \textbf{56.9} & \textbf{66.4} & \textbf{55.2} & \textbf{60.3} & \textbf{64.8} & \textbf{53.6} & \textbf{58.7} \\
    \bottomrule
    \end{tabular}
    }
    \caption{Criterion-level gold-rubric overlap for the GPT-4o backbone, averaged over five runs. Precision, recall, and F1 are reported as percentages. SkillOpt keeps the rubric generator fixed; SC-CoEvo denotes Score-Coupled Co-Evolution.}
    \label{tab:agreement}
\end{table}

Final gold-rubric scores measure response quality but do not show whether the evolved rubric-generator skill produces more faithful criteria. Table~\ref{tab:agreement} therefore reports criterion-level overlap between final generated rubrics and gold rubrics held out from optimization. We use a separate, fixed GPT-5.5 matcher~\citep{gpt55}, also excluded from optimization, to align criteria expressing the same behavioral requirement, and macro-average question-level precision, recall, and F1. On a balanced, stratified set of 150 criterion pairs, the matcher reaches 91.3\% balanced accuracy, 0.91 macro-F1, and Cohen's $\kappa=0.83$ against adjudicated human labels; human agreement is $\kappa=0.87$~\citep{cohen1960}. Further matching and calibration details are in the supplementary material.

DecoEvo raises F1 over SkillOpt by roughly 12 points on every source benchmark, with both precision and recall increasing. This joint gain is inconsistent with merely generating more criteria, which would tend to reduce precision. SC-CoEvo instead lowers F1 by 5.3--6.3 points, consistent with score-coupled generator updates potentially discarding difficult distinctions. Together, the alignment and dynamics analyses support improved evaluation coverage as a plausible mechanism behind the final gains, rather than a more permissive rubric.

\section{Conclusion}

We introduced DecoEvo, which co-evolves solver and rubric-generator skills while decoupling generator updates from aggregate solver score. Across five benchmarks and three LLM backbones, DecoEvo achieves the highest mean in all 15 settings, including two within-domain cross-benchmark transfers. Together, these results support score decoupling as a practical design principle for evolving evaluation artifacts alongside black-box LLM skills. Our evidence is limited to settings where optimization roles share a backbone and transfer remains within medicine or writing, motivating studies with different backbones across roles, cross-domain tasks, and broader human evaluation.

\clearpage
\bibliographystyle{conference}
\bibliography{conference}

\end{document}